# DEFECT MITIGATION FOR ROBOT ARM-BASED ADDITIVE MANUFACTURING UTILIZING INTELLIGENT CONTROL AND IOT

**Matsive Ali, Blake Gassen & Sen Liu***

Department of Mechanical Engineering, University of Louisiana at Lafayette, Lafayette, LA 70503, USA

## ABSTRACT

*This paper presents an integrated robotic fused deposition modeling additive manufacturing system featuring closed-loop thermal control and intelligent in-situ defect correction using a 6-degree of freedom robotic arm and an Oak-D camera. The robot arm end effector was modified to mount an E3D hotend thermally regulated by an IoT microcontroller, enabling precise temperature control through real-time feedback. Filament extrusion system was synchronized with robotic motion, coordinated via ROS2, ensuring consistent deposition along complex trajectories. A vision system based on OpenCV detects layer-wise defects position, commanding autonomous re-extrusion at identified sites. Experimental validation demonstrated successful defect mitigation in printing operations. The integrated system effectively addresses challenges real-time quality assurance. Inverse kinematics were used for motion planning, while homography transformations corrected camera perspectives for accurate defect localization. The intelligent system successfully mitigated surface anomalies without interrupting the print process. By combining real-time thermal regulation, motion control, and intelligent defect detection & correction, this architecture establishes a scalable and adaptive robotic additive manufacturing framework suitable for aerospace, biomedical, and industrial applications.*

## NOMENCLATURE

| | |
|---|---|
| $\theta_i$ | Joint angle of the $i$-th joint |
| $\theta_{23}$ | Cumulative joint angle: $\theta_2 + \theta_3$ |
| $\theta_{234}$ | Cumulative joint angle: $\theta_2 + \theta_3 + \theta_4$ |
| $a_2$ | Link length upper arm |
| $a_3$ | Link length forearm |
| $d_1$ | Distance from base to first shoulder link |
| $d_6$ | Distance from wrist to end-effector |
| $p_x, p_y, p_z$ | Cartesian coordinates of end-effector position |
| $r_{ij}$ | Element of the rotation matrix in row $i$, column $j$ |

## 1. INTRODUCTION

Fused Deposition Modeling (FDM) traditionally relies on gantry-based systems, restricting deposition to flat layers. However, aerospace, biomedical, and architectural applications increasingly demand multi-axis, non-planar manufacturing. Robotic arms, offering six degrees of freedom (DOF), enable freeform deposition and complex surface fabrication [1], [2], [3].

Despite these advantages, robotic FDM introduces challenges. First-layer adhesion is critical; misalignment can propagate defects across layers. Linear motion synchronization between filament extrusion feed rate and robotic arm trajectory is non-trivial, especially on curves and slope [4],[5]. Additionally, maintaining stable extrusion temperature is essential for consistent material flow; minor thermal fluctuations can cause poor layer bonding or clogging [6]. Traditional open-loop systems are inadequate, necessitating closed-loop control.

Extrusion must also be dynamically controlled to match robotic speed variations. Integrating thermal and extrusion control within an IoT framework enhances modularity and real-time adaptability. Finally, defect detection and correction are vital for quality assurance. By incorporating a machine vision system with traditional methods, real-time defect identification and repair become feasible without manual intervention.

This paper introduces a complete robotic additive manufacturing system integrating ESP32-based thermal and extrusion control, ROS2-based motion planning, and an advanced OpenCV based detection and mitigation pipeline.

*Corresponding author: sen.liu@louisiana.edu

## 2. Literature

One of the most effective strategies for mitigating defects in 3D printing is the implementation of closed-loop systems. These systems integrate in-situ inspection and repair mechanisms, allowing for real-time monitoring and correction of defects as they occur. For instance, a study presented by Singh, M. et al. [7], demonstrates a layer-wise closed-loop approach that uses a robotic AM platform. This system employs in-situ inspection subsystems, such as 3D point cloud scans, to identify geometric deviations and categorize them as positive or negative defects. The online process correction subsystem then addresses these defects through re-planning and repair strategies. This approach has been shown to reduce defect percentages by volume from 10.7% to 1.3%, improve geometric tolerance, and enhance mechanical properties.

The integration of multisensory fusion and digital twins has emerged as a powerful tool for defect mitigation in AM. As detailed in study by Chen, L et al. [8], a multisensory fusion-based digital twin combines data from various sensors, such as acoustic, infrared, and coaxial vision cameras, to provide a comprehensive understanding of the printing process. This approach enables the identification of regions requiring material addition or removal and generates robot toolpaths and process parameters for defect correction. The use of machine learning algorithms further enhances the ability to predict location-specific quality and correct defects in real-time.

Real-time in-situ process monitoring is essential for detecting and correcting defects during the printing process. A study presented by Beckar P. et al. [9], outlines a monitoring system that uses optical consumer sensors to detect errors such as layer shifts and stopped extrusion. The system's modular structure allows for the integration of additional sensors and error detection methods, making it highly adaptable to various printing scenarios. The use of a robot-mounted sensor further enhances the system's versatility, enabling it to handle multiple printers without requiring a proportional increase in sensors.

Filament width deviation is a common issue in extrusion-based AM, particularly in the first layer. Research presented in [10] and [11] proposes methodologies for the automatic detection and isolation of filament width deviations. These methods employ deep learning models, such as instance segmentation and morphology-based approaches, to estimate filament width and detect deviations. The integration of fault detection and isolation techniques enables the identification of the origin of deviations, correlating them with parameters such as robot-nozzle speed and material flow rate.

The integration of machine learning and vision systems has been instrumental in controlling layer morphology and mitigating defects in AM. A study presented by João S. et al. [12] demonstrates the use of machine learning models for real-time analysis of printed layers, enabling automatic adjustments to print settings. The deployment of depth cameras and customized rotary mechanisms allows for close-range monitoring of the printed layer, ensuring high precision and accuracy. This approach has been particularly effective in 3D concrete printing, where precise adjustments to parameters are crucial for achieving high-quality prints [10].

The literature reveals significant progress in defect detection and correction using closed-loop control, sensor fusion, and machine learning in additive manufacturing. However, research on feedback control and real-time processing of sensor signal and sending the precise coordinates to a robot arm-based AM system to mitigate the layer defects is still underexplored. In this paper, we have re-designed a commercial 6-DOF robot arm with filament extruding hotend and controller to enable the 3D printing tasks in a large format and with freedom. The camera-based defect localization enables to send precise coordinates to a robot-mounted hotend for targeted defect mitigation based on inverse kinematics. This gap presents an opportunity to develop a novel closed-loop framework for real-time defect correction and vision sensing using a robot-arm based additive manufacturing system.

## 3. Experimental Setup

The re-designed 6-DOF robot arm-based additive manufacturing system is developed based on a commercial ViperX 300S robot arm, hotend nozzle, hotend mount, wire extruder, thermistor, and control unit, as shown in Fig. 1. The end effector is replaced with an E3D hotend including a heater cartridge, Semitec 104GT-2 thermistor, heatsink, fan, and 0.4 mm nozzle. The hotend is mounted via a custom 3D-printed mounting bracket as illustrated in Fig. 1. Filament is fed through a PTFE Bowden tube and extruded using a NEMA17 stepper motor, driven by an A4988 driver as given in Fig. 2.

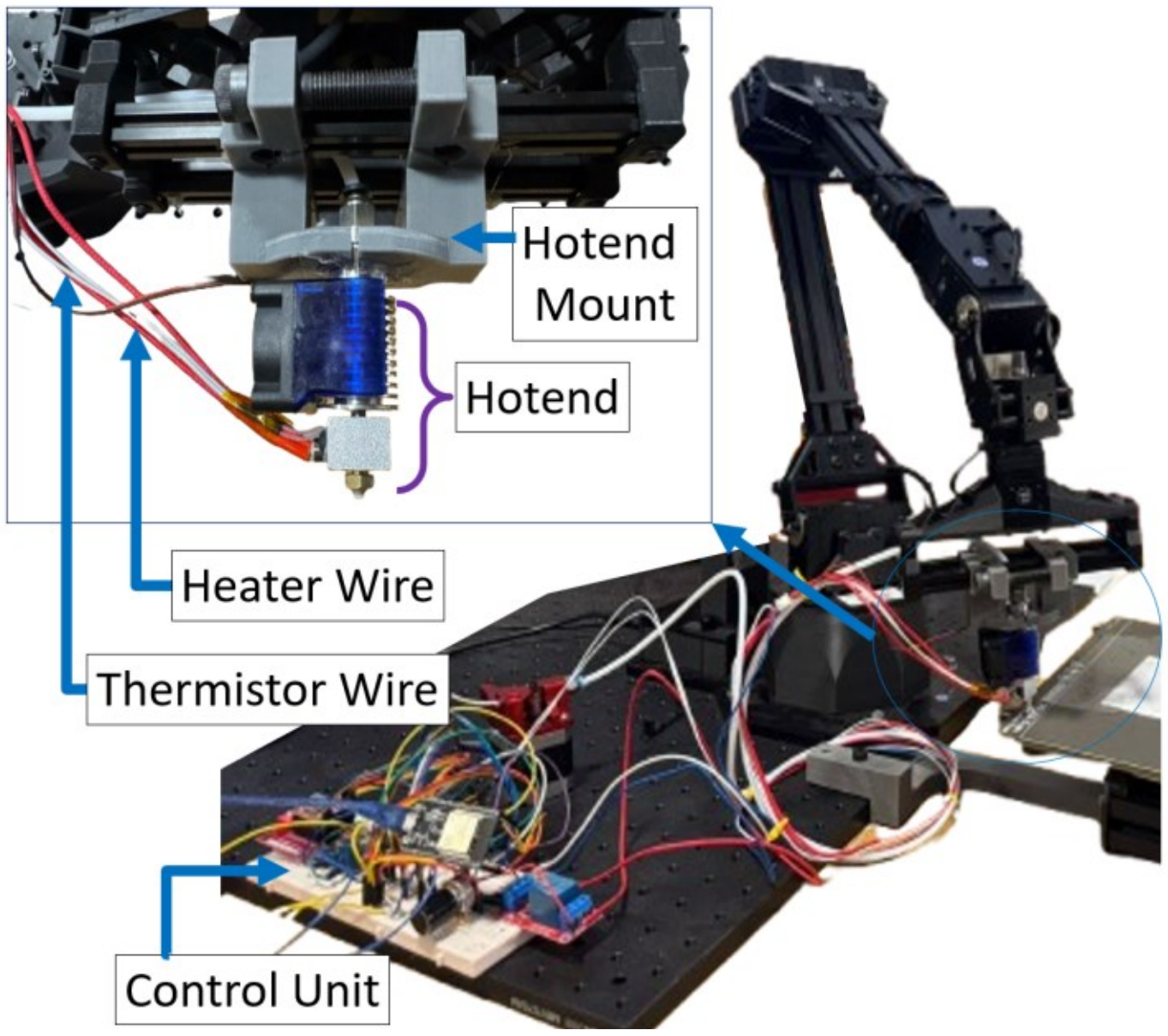

**Figure 1.** Modified Viper x300s robot arm with custom mounting bracket to attach the E3D hotend.

The heating system is powered by a 12V supply. The ESP32 microcontroller reads thermistor voltage using an analog pin and calculates the temperature via the Steinhart–Hart equation. A 10kΩ potentiometer sets the desired temperature and is scaled to a 0–250 °C range. A relay switches the heater on/off when the sensed temperature is below the setpoint as given in Fig. 2. The extruder can be activated once the set temperature is reached. Direction and step pulses are sent by the ESP32, allowing manual or programmatic control of feed length and direction.

The robotic arm is controlled using ROS2 Humble on a host Ubuntu 22.04 LTS computer. Cartesian commands are converted to joint-space trajectories using inverse kinematics (IK). The hotend end-effector is commanded to follow linear paths synchronized with extrusion. ROS2 allows modular integration with the ESP32 via serial commands and topic-based control.

In addition to the core extrusion and motion hardware, the defect detection system was integrated using the OAK-D HD Camera Development Kit and the OpenCV AI Machine Vision Kit. The OAK-D camera, equipped with depth sensing and high-resolution RGB capture, enables accurate acquisition of post-deposition surface imagery. This camera is mounted on a 3d printer bed frame with a designed mounting to hold it in place as shown in Fig. 3. Captured images are processed through an onboard pipeline and enhanced using OpenCV libraries to detect anomalies and positional deviations in the printed layer. Identified defect coordinates are subsequently mapped to the robot's reference frame, allowing precise realignment. The robotic arm can then be commanded to traverse to the detected defect location and perform localized filament re-extrusion, enabling in-process defect mitigation without interrupting the overall print workflow.

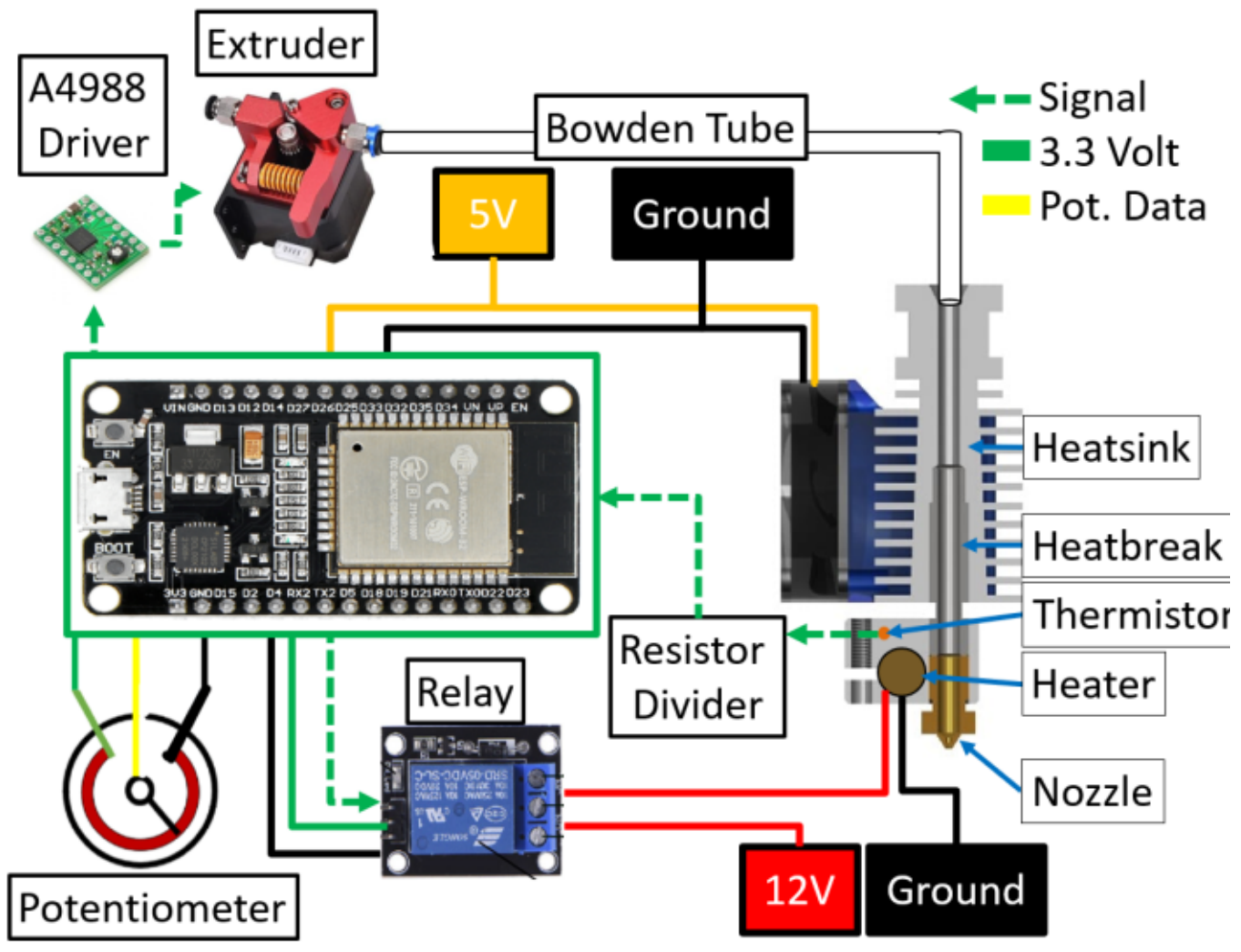

**Figure 2.** Diagram for PLA filament extrusion and temperature control setup.

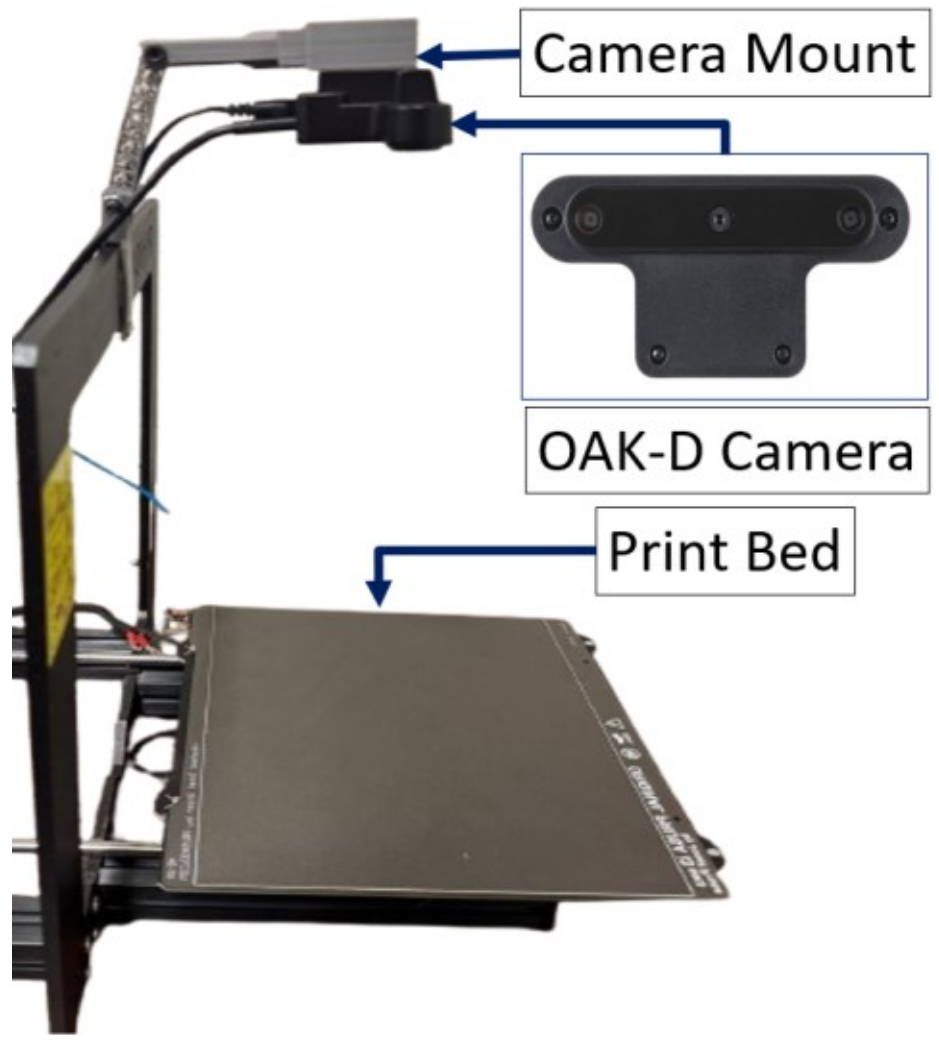

**Figure 3.** Oak-D camera attached to print bed with mount for vision system.

## 4. Methodology

### 4.1 Robot arm kinematics

The spatial configuration of the ViperX 300S robotic arm is mathematically modeled using homogeneous transformation matrices. The arm comprises five revolute joints, and the relationship between the base frame and the end-effector frame is expressed through a sequence of individual joint transformations. Each transformation accounts for the rotation and translation introduced by the corresponding joint. The overall transformation matrix $T_0^6$ from the robot base to the end-effector is obtained by sequentially multiplying the individual joint transformation matrices:

$$T_0^6 = T_0^1 \times T_1^2 \times T_2^3 \times T_3^4 \times T_4^5 \times T_5^6 \tag{1}$$

where $T_i^{i+1}$ represents the homogeneous transformation from joint $i$ to joint $i+1$. The forward kinematics model incorporates both rotational and translational components. The rotational components are functions of the joint angles $\theta_1$ through $\theta_5$, while the translational components are functions of the robot's link parameters height from base to first shoulder link, $d_1$; length of first arm link (upper arm), $a_2$; length of second arm link (forearm), $a_3$; and wrist to gripper distance write this as passage, $d_6$. The matrix elements explicitly capture the orientation and position of the end-effector in the three-dimensional workspace. The rigid body final transformation matrix $T_0^6$ is structured as:

$$T_0^6 = \begin{bmatrix} r_{11} & r_{12} & r_{13} & p_x \\ r_{21} & r_{22} & r_{23} & p_y \\ r_{31} & r_{32} & r_{33} & p_z \\ 0 & 0 & 0 & 1 \end{bmatrix} \tag{2}$$

where $r_{ij}$ represent the rotation matrix elements, and $p_x$, $p_y$, and $p_z$ represent the position of the end-effector. The terms are nonlinear functions of the joint angles and are explicitly defined based on the summation of joint variables to capture the cumulative rotations, such as $\theta_{23} = \theta_2 + \theta_3$ and $\theta_{234} = \theta_2 + \theta_3 + \theta_4$.

$$\left.\begin{aligned}
r_{11} &= cos\theta_1(cos\theta_{234}cos\theta_5 - sin\theta_{234}sin\theta_5) \\
r_{12} &= -cos\theta_1(cos\theta_{234}sin\theta_5 + sin\theta_{234}cos\theta_5) \\
r_{13} &= cos\theta_1 sin\theta_{234} \\
r_{21} &= sin\theta_1(cos\theta_{234}cos\theta_5 - sin\theta_{234}sin\theta_5) \\
r_{22} &= -sin\theta_1(cos\theta_{234}sin\theta_5 + sin\theta_{234}cos\theta_5) \\
r_{23} &= sin\theta_1 sin\theta_{234} \\
r_{31} &= sin\theta_{234}cos\theta_5 + cos\theta_{234}sin\theta_5) \\
r_{32} &= sin\theta_{234}sin\theta_5 - cos\theta_{234}cos\theta_5) \\
r_{33} &= cos\theta_{234}
\end{aligned}\right\} \quad (3)$$

$$\left.\begin{aligned}
p_x &= cos\theta_1(a_2cos\theta_2 + a_3cos\theta_{23} + d_6sin\theta_{234}) \\
p_y &= sin\theta_1(a_2cos\theta_2 + a_3cos\theta_{23} + d_6sin\theta_{234}) \\
p_Z &= d_1 + a_2sin\theta_2 + a_3sin\theta_{23} - d_6cos\theta_{234}
\end{aligned}\right\} \quad (4)$$

The inverse kinematics of the ViperX 300S robot arm is formulated through the position equations $p_x$, $p_y$, and $p_z$, which define the end-effector's coordinates *(x, y, z)* in terms of the joint $\theta_1$, $\theta_2$, $\theta_3$, and $\theta_4$. By knowing the desired end-effector position in the workspace, these equations can be inverted to calculate the corresponding joint angles required to achieve that position. Thus, $p_x$, $p_y$, and $p_z$, serve as the basis for determining the necessary joint configurations given a specified target point.

### 4.2 IoT microcontroller system

Now end-effector mounted with hotend thermistor circuit data is read through a voltage divider, and resistance is converted to temperature using calibrated coefficients of Steinhart–Hart equation for that thermistor [13]. The controller compares the current temperature to the required temperature defined by potentiometer. If the temperature falls short, a GPIO pin activates a relay to drive the heater. This bang-bang relay system (hysteresis relay system) ensures reliable thermal closed loop control.

Once the desired hotend temperature is reached, the extruder can be actively controlled. Filament extrusion is controlled via manual input or conditionally triggered logic. The controller sets direction and step pulses based on analog inputs, supporting both clockwise (extrude) and counter-clockwise (retract) modes. The extrudate rate is adjusted according to the robot velocity to optimized material flow. Various system information can be monitored using local webserver.

### 4.3 Vision system for defect detection

A high-resolution OAK-D camera is mounted rigidly above the 3D printer bed, integrated into the frame structure to provide a stable, top-down view of the print surface. After the deposition of each layer, the printing process pauses momentarily, allowing the camera to capture an image of the current layer. Due to slight perspective distortions resulting from the mounting geometry, the captured image is not perfectly aligned with the print bed. To correct this, four reference points corresponding to the corners of the print area are manually selected. A perspective transformation (homography) is then applied to warp the image into a rectified, rectangular shape of 400×400 pixel that accurately represents the print surface from a true top-down perspective as shown in Fig. 4. The indicated bounded space of 300×300 pixels in Fig. 4 is our region of interest. This rectified image is subsequently used for further analysis, such as defect detection or print quality evaluation.

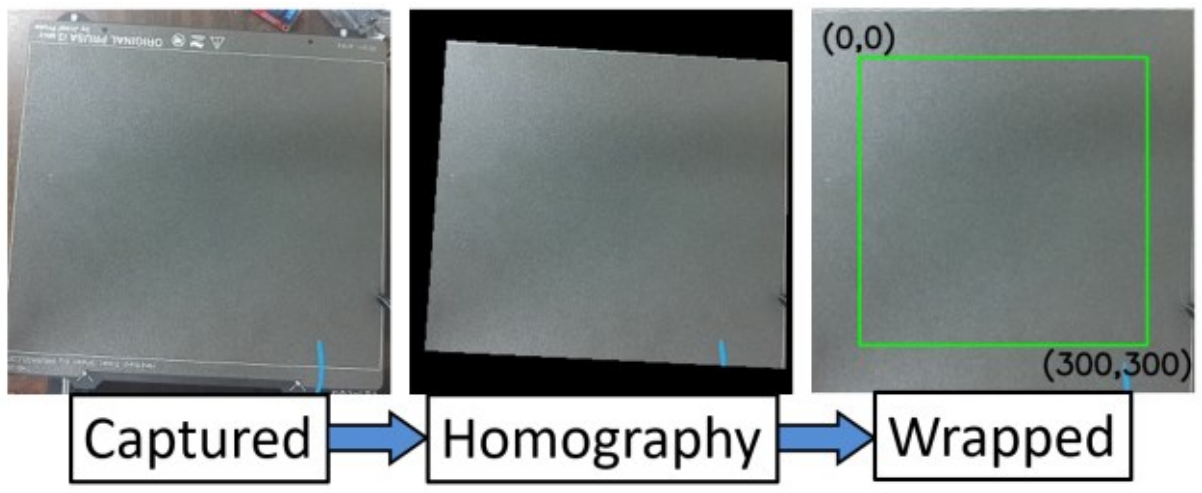

**Figure 4.** Image transformation using homography for 3D printer bed alignment.

### 4.4 End-to-End Architecture

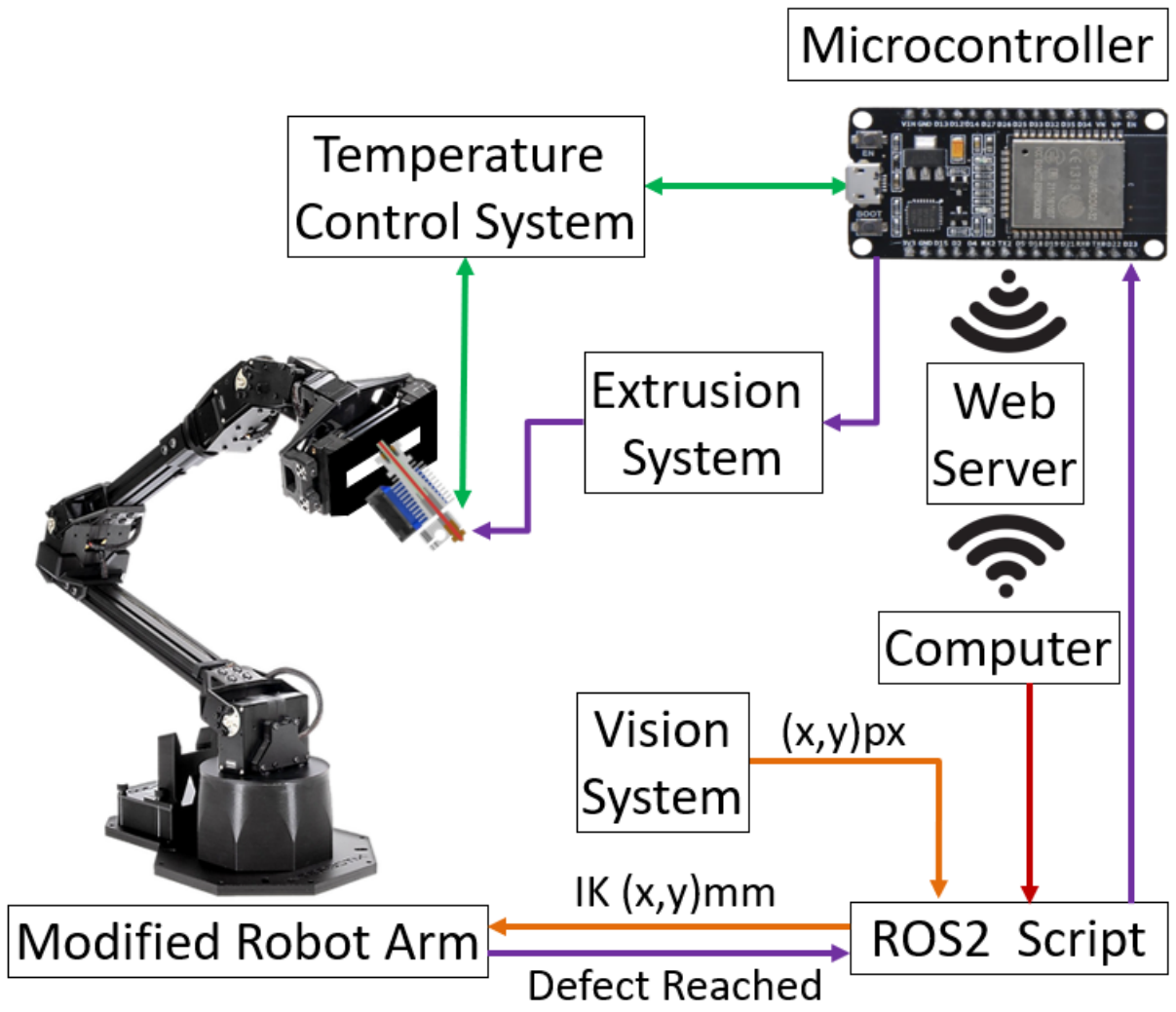

**Figure 5.** Complete system diagram featuring the modified robotic arm, temperature control unit, vision system, and extrusion mechanism for autonomous defect localization and mitigation.

The complete robotic additive manufacturing system follows a closed-loop workflow integrating thermal control, real-time monitoring, defect detection, and autonomous correction. The process begins after each layer is printed. An ESP32 microcontroller regulates hotend temperature via thermistor feedback, maintaining optimal extrusion conditions. While the OAK-D camera captures an overhead image of the surface. The image is rectified using homography to align with the print bed. OpenCV algorithms then enhance, threshold, and segment the image to identify defect regions. The centroids of these regions are mapped to Cartesian coordinates using the known camera-to-bed transformation.

ROS2 commands the robotic arm to move to each defect location, and the ESP32 triggers localized extrusion to repair them. This real-time, automated correction process eliminates the need for manual inspection, ensuring print quality. Figure 5 shows the integrated system, demonstrating coordinated control of vision, motion, and extrusion for adaptive robotic manufacturing. The ESP32 uploads data to the local webserver that can be accessed using a computer to monitor current operation.

## 5. RESULTS AND DISCUSSION

We created a rectangular sample of ($100 \times 100$ mm$^2$) with 49 defects of 2 mm holes. A Python script utilizing OpenCV is employed to process the captured images for defect detection. Initially, the image undergoes contrast enhancement to improve the visibility of features and ensure consistent brightness across the frame. Following enhancement, thresholding is applied to binarize the image, separating the printed structure from the background. The binarized image is then subjected to segmentation, isolating regions of interest where potential defects may exist. Finally, quantitative analysis is performed on the segmented regions to determine the position, size, and characteristics of any detected anomalies on the printed layer.

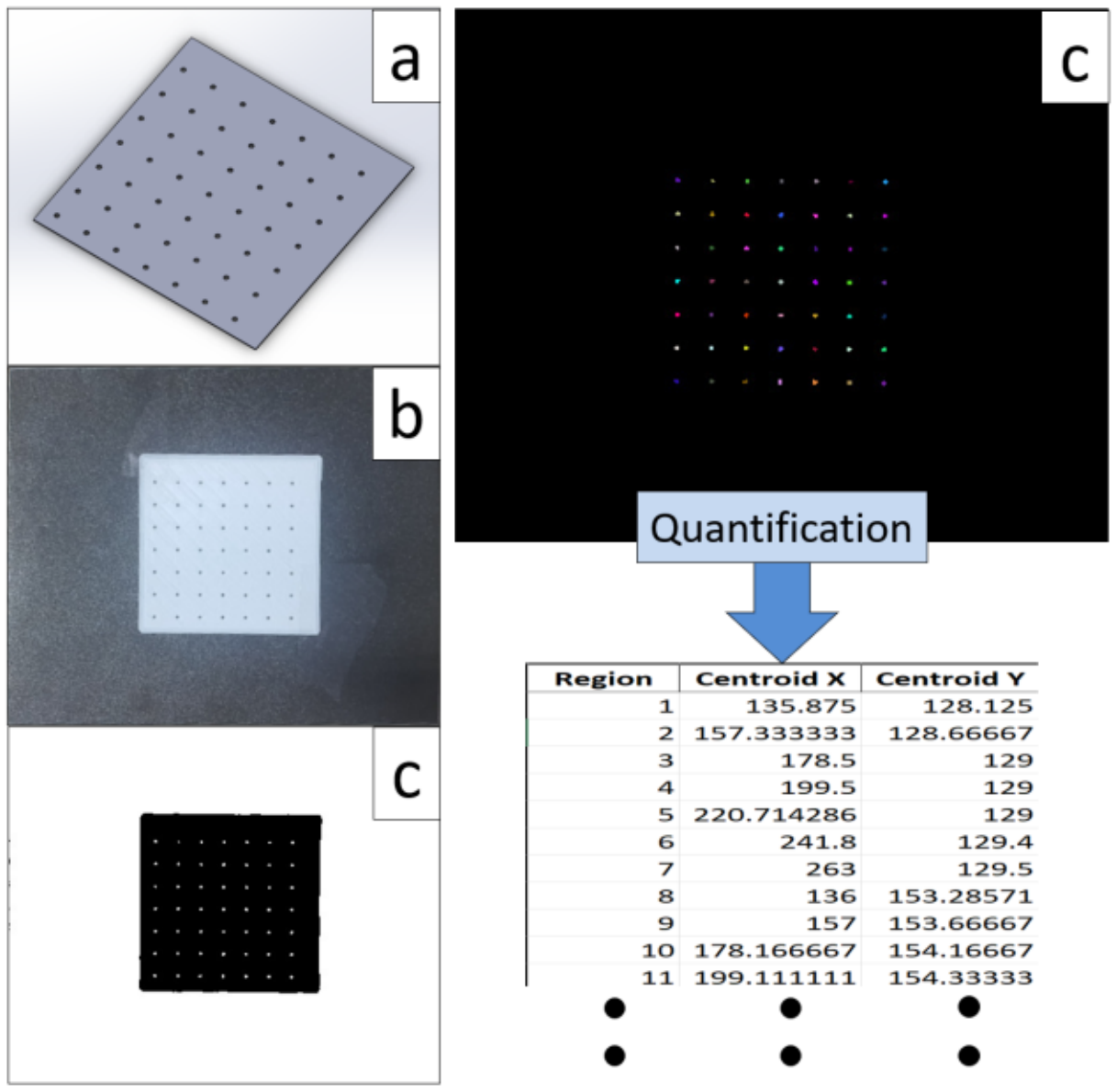

**Figure 6.** (a) Designed sample with defect of 2mm, (b) image of sample printed on bed, (c) thresholding of image, (d) segmentation and quantification of image to get position of each defect region.

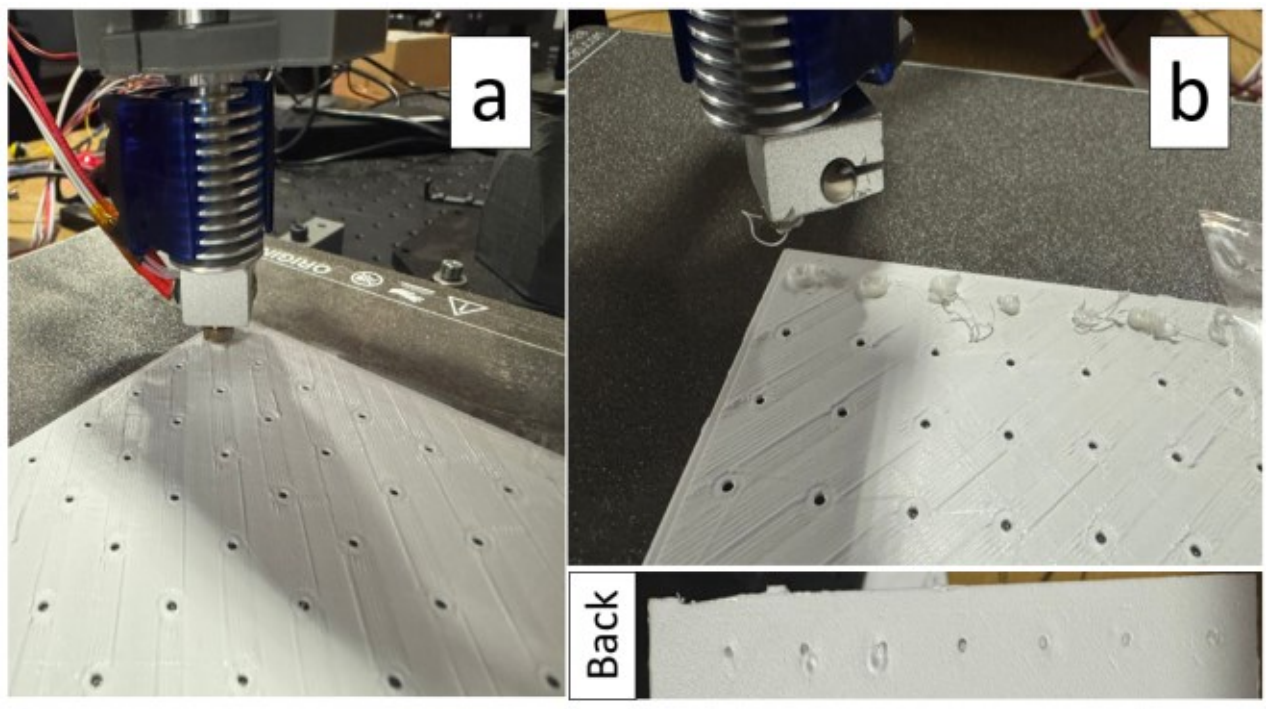

**Figure 7.** (a) Robot arm reached defect location (b) System fixed first column of 7 defects.

Once a defect is detected, its centroid position is translated into a Cartesian pose of the robot arm. Thus, robot is able to move to all 49 defect locations. Then ROS2 commands the robot to move to defect location. The ESP32 activates the extruder to deposit filament locally, filling in the defect. This process enables closed-loop correction and significantly reduces failure rates. The system moves toward autonomy by identifying and fixing faults before the next layer is deposited.

The current system was tested using surface-level defects with controlled geometry (e.g., 2 mm circular voids), representing visible surface anomalies such as under-extrusion or missing filament paths. All 49 defects were correctly detected and mitigated, indicating a 100% detection success rate within the scope of these test conditions. A limitation of the current vision setup is that it may fail to reliably detect defects smaller than approximately 0.67 mm in diameter, which corresponds to about one pixel in the rectified image. However, the system does not currently detect subsurface defects such as internal porosity or lack of fusion, which require volumetric sensing (e.g., ultrasound or X-ray). The method is currently optimized for detecting visible top-layer discontinuities using 2D vision, such that each printed layer is defect-free before proceeding to the next layer.

This paper demonstrates a 6-DOF robotic arm-based 3D printing system with closed-loop temperature regulation, extrusion control, and defect correction. The combination of ESP32-based thermal control, ROS2 motion planning, and Python vision feedback forms a modular architecture for intelligent additive manufacturing.

While the proposed system demonstrates effective defect detection and correction on planar surfaces with uniform defects, it currently assumes minimal surface curvature and regular feature geometry. This simplifies vision alignment and Cartesian-to-joint space transformation. Future work involves addressing applications involving irregular geometries or curved surfaces, challenges such as occlusions, variable lighting, and nonlinear surface mapping would arise. Addressing these would require more advanced vision algorithms, real-time depth fusion, and adaptive kinematic path planning of the 6-DOF robot arm along with machine learning. These scenarios represent a significant next step in extending the system's applicability.

In the direction of future research, we would be integrating PID and RL-based extrusion control to fix over-extrusion of filament while mitigation, expanding the defect detection dataset, and incorporating CAD-aware defect classification into the vision pipeline. This enhancement will enable the system to differentiate between intentional geometric features (e.g., designed holes) and unintended defects. For more complex real-world scenarios, we also plan to integrate higher-resolution cameras fused with depth sensing systems (e.g., LiDAR) and CAD/toolpath data to detect and classify various surface anomalies such as gaps, missing layers, over-extrusion, under-extrusion, or unintended voids. The modular design allows scaling to more advanced applications in aerospace repair, off-world printing, and adaptive manufacturing.

## 6. CONCLUSION

This work presents an integrated robotic additive manufacturing system that combines closed-loop thermal control, synchronized extrusion, and intelligent defect correction. The use of ROS2, ESP32, and OpenCV enables real-time coordination between vision, motion, and extrusion subsystems. Experimental validation confirms that the system successfully identifies and repairs defects autonomously during the print process. This architecture lays the foundation for scalable, adaptive, and intelligent 3D printing suitable for advanced manufacturing domains.

## ACKNOWLEDGEMENTS

The authors would like to express their sincere gratitude to the Board of Regents Support Fund (BoRSF) Research Competitiveness Subprogram LEQSF(2024-27)-RD-A-32, and Louisiana Transportation Research Center (LTRC) TIRE program 25-2TIRE.

## REFERENCES

[1] Sutjipto, S., Tish, D., Paul, G., Vidal-Calleja, T., and Schork, T., 2019, "Towards Visual Feedback Loops for Robot-Controlled Additive Manufacturing," *Robotic Fabrication in Architecture, Art and Design 2018: Foreword by Sigrid Brell-Çokcan and Johannes Braumann, Association for Robots in Architecture*, Springer, pp. 85–97. https://doi.org/10.1007/978-3-319-92294-2_7.

[2] Yao, Y., Zhang, Y., Aburaia, M., and Lackner, M., 2021, "3D Printing of Objects with Continuous Spatial Paths by a Multi-Axis Robotic FFF Platform," Appl. Sci., **11**(11), p. 4825. https://doi.org/10.3390/app11114825.

[3] De Backer, W., Bergs, A. P., and Van Tooren, M. J., 2018, "Multi-Axis Multi-Material Fused Filament Fabrication with Continuous Fiber Reinforcement," *2018 AIAA/ASCE/AHS/ASC Structures, Structural Dynamics, and Materials Conference*, p. 91. https://doi.org/10.2514/6.2018-0091.

[4] Kaji, F., Jinoop, A. N., Zimny, M., Frikel, G., Tam, K., and Toyserkani, E., 2022, "Process Planning for Additive Manufacturing of Geometries with Variable Overhang Angles Using a Robotic Laser Directed Energy Deposition System," Addit. Manuf. Lett., **2**, p. 100035. https://doi.org/10.1016/j.addlet.2022.100035.

[5] Pollák, M., Kočiško, M., Grozav, S. D., Ceclan, V., and Bogdan, A. D., 2024, "Suitability of UR5 Robot for Robotic 3D Printing," Appl. Sci., **14**(21), p. 9845. https://doi.org/10.3390/app14219845.

[6] Urhal, P., Weightman, A., Diver, C., and Bartolo, P., 2019, "Robot Assisted Additive Manufacturing: A Review," Robot. Comput. Integr. Manuf., **59**, pp. 335–345. https://doi.org/10.1016/j.rcim.2019.05.005.

[7] Singh, M., Ruan, F., Xu, A., Wu, Y., Rungta, A., Wang, L., Song, K., Choset, H., and Li, L., 2023, "Toward Closed-Loop Additive Manufacturing: Paradigm Shift in Fabrication, Inspection, and Repair," *2023 IEEE/RSJ International Conference on Intelligent Robots and Systems (IROS)*, IEEE, pp. 10066–10073.

[8] Chen, L., Yao, X., Liu, K., Tan, C., and Moon, S. K., 2023, "Multisensor Fusion-Based Digital Twin in Additive Manufacturing for in-Situ Quality Monitoring and Defect Correction," Proc. Des. Soc., **3**, pp. 2755–2764.

[9] Becker, P., Spielbauer, N., Roennau, A., and Dillmann, R., 2020, "Real-Time in-Situ Process Error Detection in Additive Manufacturing," *2020 Fourth IEEE International Conference on Robotic Computing (IRC)*, IEEE, pp. 426–427.

[10] Yang, X., Lakhal, O., Belarouci, A., and Merzouki, R., 2023, "Automatic Detection and Isolation of Filament Width Deviation during 3-D Printing of Recycled Construction Material," IEEE/ASME Trans. Mechatronics, **29**(3), pp. 1939–1948.

[11] Yang, X., Lakhal, O., Belarouci, A., Youcef-Toumi, K., and Merzouki, R., 2023, "Experimental Workflow Implementation for Automatic Detection of Filament Deviation in 3D Robotic Printing Process," *2023 IEEE International Conference on Robotics and Automation (ICRA)*, IEEE, pp. 12309–12315.

[12] Silva, J. M., Wagner, G., Silva, R., Morais, A., Ribeiro, J., Mould, S., Figueiredo, B., Nóbrega, J. M., and Cruz, P. J. S., 2024, "Real-Time Precision in 3D Concrete Printing: Controlling Layer Morphology via Machine Vision and Learning Algorithms," Inventions, **9**(4), p. 80.

[13] Steinhart, J. S., and Hart, S. R., 1968, "Calibration Curves for Thermistors," *Deep Sea Research and Oceanographic Abstracts*, Elsevier, pp. 497–503.